\documentclass[conference]{IEEEtran}
\IEEEoverridecommandlockouts
\usepackage{cite}
\usepackage{amsmath,amssymb,amsfonts}
\usepackage{algorithmic}
\usepackage{graphicx}
\usepackage{textcomp}
\usepackage{xcolor}
\usepackage{booktabs} 
\usepackage{multirow}
\usepackage{hyperref}
\usepackage{url}
\usepackage{siunitx}
\usepackage{caption,subcaption}
\usepackage[acronym]{glossaries}
\usepackage{tabularx, booktabs, multirow}
\usepackage{ragged2e}
\usepackage{makecell}

\usepackage{pgfplots}
\pgfplotsset{compat=1.18}

\def\BibTeX{{\rm B\kern-.05em{\sc i\kern-.025em b}\kern-.08em
    T\kern-.1667em\lower.7ex\hbox{E}\kern-.125emX}}

\newacronym{ba}{BA}{Biological Age}
\newacronym{ca}{CA}{Chronological Age}
\newacronym{hpp}{HPP}{Human Phenotype Project}
\newacronym{ml}{ML}{Machine Learning}
\newacronym{lgbm}{LightGBM}{Light Gradient Boosting Machine}
\newacronym{shap}{SHAP}{SHapley Additive exPlanations}
\newacronym{mlp}{MLP}{Multi-Layer Perceptron}
\newacronym{imt}{IMT}{Intima-Media Thickness}
\newacronym{bp}{BP}{Blood Pressure}
\newacronym{bmi}{BMI}{Body Mass Index}
\newacronym{hba1c}{HbA1c}{Hemoglobin A1c}

\makeglossaries

\begin{document}

\title{How Effectively Can Large Language Models Connect SNP Variants and ECG Phenotypes for Cardiovascular Risk Prediction?}

\author{Niranjana Arun Menon$^1$ , Iqra Farooq$^2$ , Yulong Li$^3$ , Sara Ahmed$^2$ , Yutong Xie$^3$, \\ Muhammad Awais$^2$ , Imran Razzak$^3$  \\
\textit{$^1$University of New South Wales, Australia} \\
\textit{$^2$University of Surrey, United Kingdom}\\
\textit{$^3$Mohamed bin Zayed University of Artificial Intelligence, UAE}
}

\maketitle

\begin{abstract}

Cardiovascular disease (CVD) prediction remains a tremendous challenge due to its multifactorial etiology and global burden of morbidity and mortality. Despite the growing availability of genomic and electrophysiological data, extracting biologically meaningful insights from such high-dimensional, noisy, and sparsely annotated datasets remains a non-trivial task. Recently, LLMs has been applied effectively to predict structural variations in biological sequences. In this work, we explore the potential of fine-tuned LLMs to predict cardiac diseases and SNPs potentially leading to CVD risk using genetic markers derived from high-throughput genomic profiling. We investigate the effect of genetic patterns associated with cardiac conditions and evaluate how LLMs can learn latent biological relationships from structured and semi-structured genomic data obtained by mapping genetic aspects that are inherited from the family tree. By framing the problem as a Chain of Thought (CoT) reasoning task, the models are prompted to generate disease labels and articulate informed clinical deductions across diverse patient profiles and phenotypes. The findings highlight the promise of LLMs in contributing to early detection, risk assessment, and ultimately, the advancement of personalized medicine in cardiac care.

\end{abstract}


\section{Introduction}
Cardiovascular disease (CVDs) remain the most common cause of mortality globally.  In 2023, approximately 20.5 million people died from CVDs~\cite{Pcr_2023}, accounting for about one-third of all global deaths. This marks a significant increase from the estimated 17.9 million CVD deaths in 2019. Early identification of high-risk patients and timely initiation of appropriate treatment are crucial in mitigating adverse health outcomes associated with CVD. Genome-wide association studies (GWAS)~\cite{noauthor_genomewise_2025} has uncovered numerous single nucleotide polymorphisms (SNPs) linked to CVDs, many of which are located in non-coding regions of the genome. These genetic variations can influence gene activity by impacting transcription factor (TF) binding sites, potentially resulting in changes that affect cardiovascular traits or disease development. Gaining insight into which variants are involved and their impact on TF binding is essential for uncovering the molecular basis of these conditions. While inherited genetic variation has long been established as a foundational determinant of cardiovascular disease (CVD) susceptibility, emerging research such as the integrative multi-omics framework proposed by Nam et al.\cite{nam_harnessing_2024} and the systems-level phenotyping approach outlined by Yurkovich et al.\cite{yurkovich_transition_2024} emphasizes the critical role of combining genomic data with dynamic phenotypic biomarkers, including electrocardiographic (ECG) features. Such integrative strategies are increasingly recognized for their potential to refine risk stratification, elucidate disease mechanisms, and enable more precise predictive modeling in complex trait disorders like CVD. Single nucleotide polymorphisms (SNPs) have been widely studied in relation to arrhythmias, myocardial infarction, and sudden cardiac death, yet the mechanistic links between these variants and electrophysiological manifestations remain poorly understood~\cite{tamariz_systematic_2019}. Understanding how specific variants impact electrophysiological processes could enhance risk stratification in arrhythmic patients. 

\begin{figure}
    \centering
    \includegraphics[width=0.98\linewidth]{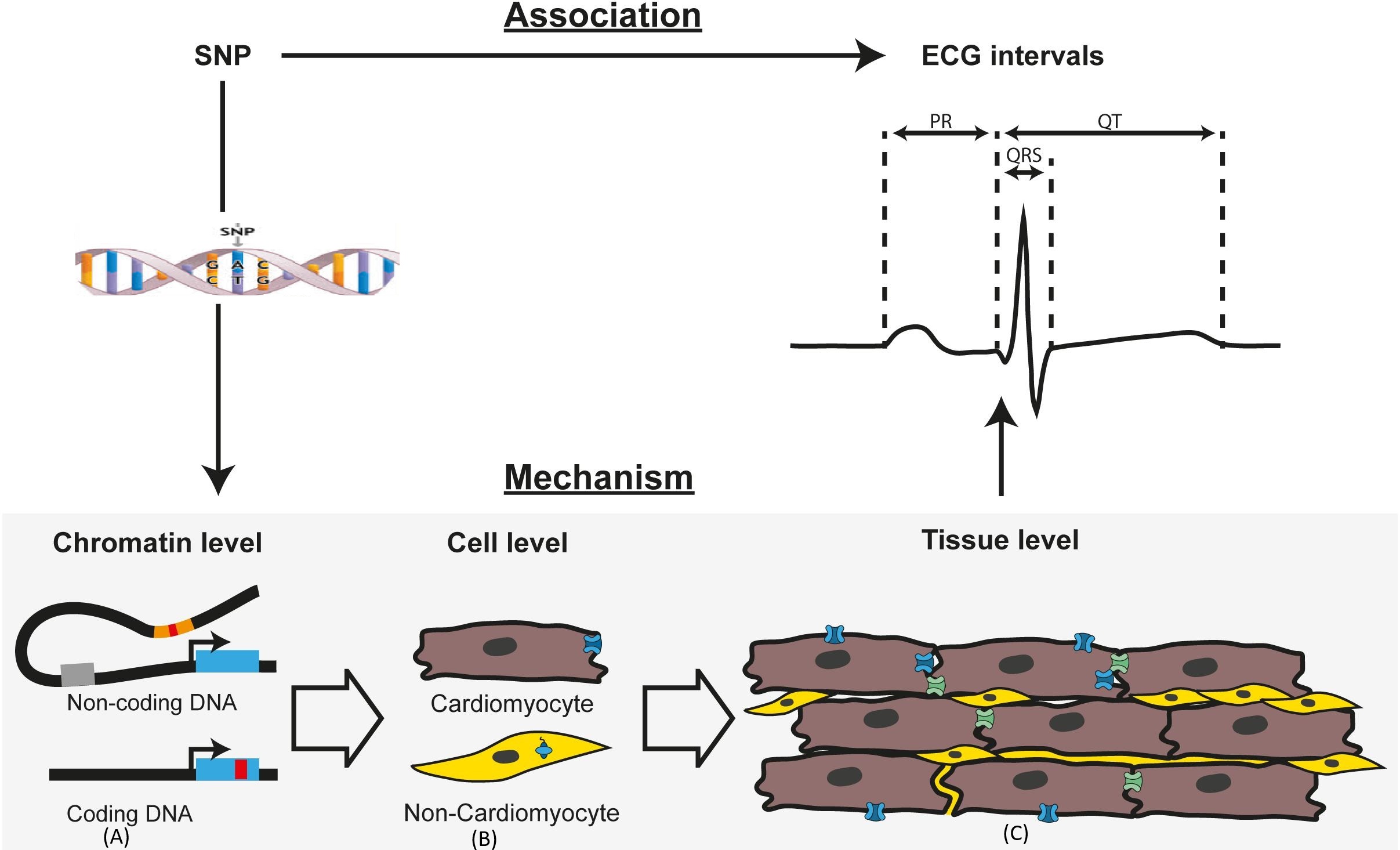}
    \caption{SNPs influence ECG intervals (PR, QRS, QT) across biological levels: (A) Chromatin—identify genes affected by coding or non-coding SNPs; (B) Cell—determine cell types expressing the gene; (C) Tissue—link gene expression to tissue electrophysiology and ECG phenotypes.}
    \label{fig:snps}
\end{figure}

Deciphering these relationships requires modeling complex interactions between genomic features and time-series cardiovascular ECG data. Recently, LLMs have shown transformative potential across natural language processing tasks and are now being increasingly explored in scientific domains for modeling structured and unstructured data. 
Recent work has also demonstrated that LLMs are capable of high-fidelity extraction of clinically relevant information from free-text electronic health records (EHRs). Gu et al.~\cite{gu_scalable_2025} applied zero-shot and few-shot prompting with general-purpose LLMs such as GPT-3.5 and GPT-4 to automate clinical concept recognition, temporal relation extraction, and patient outcome prediction across large-scale EHR datasets. Their study emphasizes the adaptability of LLMs to heterogeneous medical domains, even in the absence of task-specific fine-tuning. Although competitive with rule-based and supervised baselines, their findings also highlight key limitations such as hallucination risks and sensitivity to prompt phrasing. These insights motivate our inclusion of refined prompt engineering strategies and validation measures when using LLMs for structured genomic and ECG reasoning.
\newline
In genomics, traditional statistical models such as GWAS or polygenic risk scores often fall short in capturing nonlinear relationships and epistatic interactions across the genome \cite{li_modeling_2025}. LLMs, by contrast, provide a scalable mechanism to encode long-range dependencies and positional context in biological sequences, potentially allowing for the discovery of complex genotype-phenotype relationships that would otherwise remain obscured. Leveraging LLMs for multimodal integration of SNP data and ECG-derived features offers a novel, interpretable framework for risk stratification and biomarker discovery. Moreover, advances in multimodal learning allow LLMs to uncover clinically relevant cross-modal patterns, making them well-suited for tasks such as CVD risk classification, variant prioritization, and waveform anomaly detection~\cite{srinivasan_applications_2025}\cite{cakal_cardiovascular_2023}. Despite progress in genetic and cardiac modeling separately, studies integrating both within a unified, explainable architecture remain sparse, which highlights an urgent need for comprehensive, interpretable, and scalable cardiogenomic models.


Cardiovascular phenotyping, particularly through electrocardiogram (ECG) analysis, adds another dimension of complexity. ECG waveforms are temporally rich biosignals that reflect a wide spectrum of physiological and pathological cardiac states. When interpreted through neural architectures such as transformer-based models, ECG signals can reveal subclinical patterns of disease \cite{degroat_discovering_2024}. We explore the genomic underpinnings of cardiac conditions by analyzing genetic patterns that may contribute to disease susceptibility and progression. Specifically, our study evaluates the ability of large language models (LLMs) to uncover latent biological relationships embedded within both structured (e.g., variant call formats, annotated gene panels) and semi-structured genomic data (e.g., clinical notes, pedigree information). We explore different tokenization strategies for genetic markers (e.g., SNP IDs, genomic loci), data embeddings for biosignal-derived features, and prompt engineering for multimodal learning within transformer-based architectures. In addition to mapping gene-disease associations, we incorporate family-based genetic information to identify potential hereditary links and variant propagation patterns for each patient. 
To systematically evaluate model performance, we formulate the analysis as a series of classification tasks—ranging from predicting the presence of specific cardiac phenotypes to assessing patient-specific risk profiles based on inherited genetic variants. The results demonstrate that LLMs, when fine-tuned on multi-modal genomic data, can extract high-order latent representations from biological and physiological data streams, facilitating improved diagnostic performance and interpretability. Besides, it can effectively generalize across diverse patient cohorts and phenotypic presentations. This could democratize access to genomic diagnostics, where interpretability and computational tractability are often limiting factors. To this end, we have the following key contributions






\begin{itemize}
    \item Construct a harmonized cardiogenomic dataset by aligning high-resolution SNP genotyping data with ECG-derived morphological and temporal features extracted from the HPP, enabling joint modeling of genetic and electrophysiological profiles.
    \item Fine-tune open-source large language models on tokenized biological sequences (SNP k-mers, haplotypes) and dense ECG signal embeddings using modality-aware positional encodings, enabling unified representation learning across disparate biomedical data types.
    \item Conduct in-depth interpretability analyses by evaluating self-attention distributions and gradient-based feature attribution scores to uncover genomic loci and waveform segments predictive of cardiovascular disease risk, facilitating biologically meaningful insights.
    \item Demonstrate the feasibility of LLMs as scalable, plug-and-play architectures for integrative modeling in cardiogenomics, providing a pathway toward interpretable multimodal disease prediction and biomarker discovery across genomic and physiological data sources.
\end{itemize}


\begin{figure*}[h]
    \centering
    \includegraphics[width=0.960\textwidth]{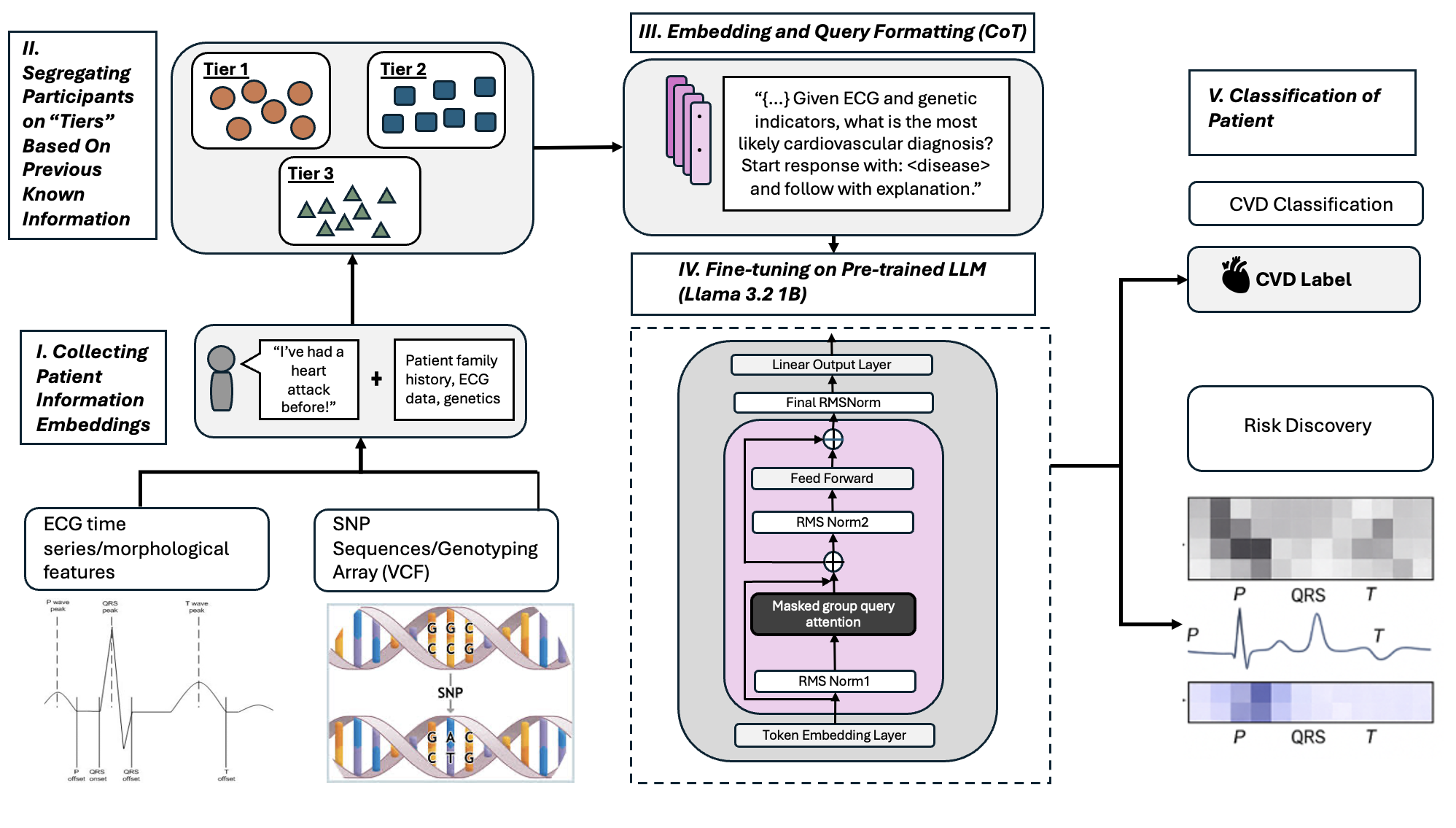}
    \caption{Integrative Cardiogenomics Framework: Linking SNP Variants and ECG Phenotypes for Explainable CVD Prediction}
    \label{fig:example-generation}
\end{figure*}

\section{Data Sources and Integration}
Figure~\ref{fig:snps} illustrates the relationship between gene expression in cardiac tissues and electrophysiological phenotypes derived from electrocardiogram (ECG) signals. To investigate this relationship, we conducted experiments on a harmonized cardiogenomic dataset that integrates high-resolution single nucleotide polymorphism (SNP) genotyping with morphological and temporal ECG features, curated from the PhenoAI HPP repository of datasets. This repository consists of cohort data from individuals residing in the Asia/Jerusalem time zone, meaning that the population is not geographically stratified across multiple time zones. As such, temporal or environmental confounding factors linked to circadian rhythm variation across regions are minimized. This unified multimodal dataset provides a foundation for jointly modeling genetic variation and electrophysiological signatures, facilitating more effective cardiovascular risk stratification. 

To ensure robust preprocessing and alignment across modalities, we transformed the complex datasets into the following formats:
    \begin{itemize}
    \item \textbf{Genotype and ECG data:} Each participant’s SNP variants and ECG metrics were subjected to variants quality control (QC), including filters for minor allele frequency (MAF $< 0.01$), Hardy-Weinberg equilibrium (HWE), and genotype posterior (GP) thresholds. The resulting data was stored in \texttt{.json} format to enable efficient retrieval and per-participant processing.
    \item \textbf{Medical Conditions:} Diagnostic cardiac condition annotations (obtained through previous known medical history) for each participant was stored in \texttt{.csv} format, providing phenotypic ground truth.
\end{itemize}

The heterogeneous datasets were linked through unique participant identifiers to ensure proper alignment across genomic, electrophysiological, and clinical modalities. To enable coherent learning from this multimodal input, we fine-tuned open-source large language models (LLMs) on tokenized biological sequences and dense ECG-derived embeddings in a unified representation space. Our training setup ensured to segregate the participants into three self-defined tiers, each reflecting varying levels of label availability; from fully supervised genotype-phenotype associations to cases requiring latent inference based on clustered ECG morphology alone. This tiered approach allowed us to evaluate model generalization across a gradient of clinical supervision and biological abstraction.


Fine-tuning was performed using Low-Rank Adaptation (LoRA)~\cite{hu_2021_lora}, applied selectively to attention and MLP layers of the transformer architecture, with the hyperparameters of \texttt{rank = 8} and \texttt{alpha = 16}. Training was conducted on an AWS EC2 instance equipped with NVIDIA A10 GPUs and used mixed precision (fp16) to reduce memory usage and improve training efficiency. Model checkpoints were saved every 50 steps, and early stopping was employed based on best validation accuracy. The fine-tuning process and architecture are summarized in Figure\ref{fig:example-generation}.

\section{Feature Engineering and Pipeline Overview}
\subsection{Data Format and Preprocessing}

Our dataset includes $8,856$ participants, with multimodal data in the form of \texttt{.json} files containing ECG and SNP features, and a \texttt{.csv} file of clinical condition labels. Due to the scarcity of cardiac conditions as labelled data (only $350$ participants with confirmed diagnoses), we stratified the entire cohort into three tiers to support downstream modeling strategies:

\begin{itemize}
    \item \textbf{Tier 1:} Consists of participants with well-established cardiac diagnoses.
    \item \textbf{Tier 2:} Consists of participants exhibiting indirect cardiac indicators.
    \item \textbf{Tier 3:} Unlabeled Participants or Participants with no known prior cardiac diagnosis.
\end{itemize}

We utilize clinical keywords to assign participants to appropriate tiers. For instance, Tier 1 includes individuals explicitly diagnosed with high-risk conditions such as \texttt{atrial fibrillation}, identified through similarity scores computed via a BioBERT model that was previously trained on medical terminology~\cite{deka2022evidence}. Tier 2 captures participants with lower-risk yet relevant associations such as \texttt{hypertension}, indicating potential future cardiovascular complications. This tiered classification enables targeted feature engineering and modeling strategies adapted to the varying clinical significance of each subgroup. This tiered classification enables downstream feature engineering and modeling strategies tailored to each group.

\subsection{Tier 1: High-Confidence Cardiac Diagnoses}

Tier 1 encompasses participants with clinically confirmed cardiac conditions such as myocarditis, coronary atherosclerosis, or arrhythmias. To identify relevant genetic variants per participant, we first extracted condition-specific labels filtered to Tier 1 diagnoses. For each label, we retrieved corresponding curated genome-wide association studies (GWAS) and expert-reviewed SNP lists, focusing on variants with established associations to the given condition. Specifically, the catalogs were taken from the ebi GWAS repository~\footnote{\url{https://www.ebi.ac.uk/gwas/}} and electing SNPs that showed genome-wide significant associations with a $p \leq 5 \times 10^{-8}$ with the condition. This manual curation step ensures that only biologically relevant and high-confidence SNPs (identified by rsIDs) are selected, providing a consistent and interpretable genetic signature for each individual.

These condition-specific SNP sets serve two key purposes in our modeling pipeline. First, they ground the participant’s genotypes in a clinically meaningful context, helping to reduce noise introduced by irrelevant variants. Second, they function as structured priors for downstream large language models (LLMs), which are not inherently pre-trained on genetic data. By injecting disease-associated variants into the prompt construction, particularly when designing chain-of-thought reasoning by guiding the LLM to focus on the genotype-condition based knowledge.

\subsection{Tier 2: Indirect Cardiac Risk}

Tier 2 encompasses individuals with comorbidities or phenotypes that exhibit indirect but statistically supported genetic associations with cardiovascular diseases (such as hypertension or hyperlipidemia). These phenotypes are frequently considered precursors or modifiers in cardiovascular risk stratification frameworks. To systematically encode genotype information in this tier, we adopt a TF-IDF (Term Frequency–Inverse Document Frequency) representation scheme over SNP profiles, building on the framework proposed in~\cite{10876089}.

In this formulation, we model each participant’s set of SNPs as a sparse text-like vector, where each rsID is treated analogously to a token in a corpus. The term frequency (TF) component corresponds to the presence (or dosage) of a given SNP in a participant’s genotype, while the inverse document frequency (IDF) penalizes SNPs that are ubiquitous across the cohort, thereby enhancing the signal of rare or cohort-specific variants. Importantly, we preprocess SNPs by filtering for known variant-disease associations using publicly available GWAS summary stats taken from the EBI GWAS catalog repository~\cite{ebiGWASCatalog} and disease-specific SNP lists, ensuring that the vocabulary of rsIDs reflects curated domain knowledge.

The resulting TF-IDF vectors capture relative informativeness of variants across the population, and are subsequently projected into a shared feature space alongside ECG-derived embeddings and structured clinical metadata. This unified multimodal input is then passed into our fine-tuned large language model (LLM) pipeline, allowing the model to reason over intermediate genetic risk in a structured yet flexible fashion. By embedding TF-IDF representations alongside phenotypic features, the model can learn higher-order interactions.

\subsection{Tier 3: Unlabeled Participants}

Tier 3 comprises participants without any known cardiac diagnoses or labeled phenotypes. To extract a meaningful structure from this unlabeled cohort, we first construct TF-IDF representations over both SNP and ECG-derived features across all Tier 3 individuals. These embeddings are then subjected to unsupervised clustering, following approaches such as those outlined in~\cite{computation13060144}, to uncover latent genotype-phenotype groupings between individual metrics.

To assess clinical relevance, each resulting cluster is also analyzed post hoc by comparing the distribution of Tier 1 SNPs and ECG biomarkers within the cluster. This enables us to infer putative risk levels. For example, participants who genetically resemble Tier 1 individuals yet currently exhibit only normative ECG signals may be annotated with a "future-risk" pseudo-label in the cluster. This strategy allows for the enrichment of the training dataset with soft labels that reflect potential subclinical cardiovascular risk.
\newline
\newline
Following the integration of all three tiers, we subsequently generated chain-of-thought prompts per participant, incorporating the relations between genotypic signals to cardiac diseases to improve model interpretability and risk stratification.

\section{Model Training Pipeline}
\subsection{Chain-of-Thought Prompt Construction}

Each Chain-of-Thought (CoT) prompt was carefully constructed to incorporate comprehensive and participant-specific genomic and electrophysiological information, enabling the large language models (LLMs) to reason about cardiovascular risk in a clinically interpretable manner. This approach is motivated by prior work conducted on CoT-based LLM training in medical settings such as ~\cite{miao_chain_2024} and \cite{wei2023chainofthoughtpromptingelicitsreasoning}, which have demonstrated enhancement of an LLM’s ability to perform complex information synthesis and improves its capacity for context-aware inference in biomedical tasks using CoT prompting.

Specifically, each prompt integrates the following key elements per participant:
\begin{itemize}
    \item Morphological and temporal features extracted from participant ECG timeseries data, such as QRS duration, PR interval, QTc, and heart rate variability is flattened. These features capture electrophysiological phenotypes relevant to cardiac function.
    
    \item The Selected SNP variants associated with cardiac conditions. For Tier 1 and Tier 2 participants, these SNPs are directly matched from curated GWAS datasets or TF-IDF analysis whereas for Tier 3, cluster-specific variant signatures inferred from unsupervised grouping are included.
    
    \item Diagnostic labels or inferred risk categories for Tier 3 (such as "high risk") to provide clinical grounding for the prompt and guide the LLM's reasoning toward relevant phenotypes.
    
    \item Instructional query: A natural language question prompting the LLM to synthesize the combined ECG and genetic data and provide a cardiovascular diagnosis or risk assessment. 
    \begin{quote}
        \texttt{Conclusion: Based on the above SNPs and ECG findings, what is the participant is likely at risk for? Start your response with: I believe it is [<cardiac condition>] and follow up with an explanation.}
    \end{quote}
\end{itemize}

This design encourages the LLM to generate evidence-based, stepwise reasoning by explicitly referencing both electrophysiological signals and genetic predispositions. Importantly, the CoT format mitigates the risk of purely associative or spurious predictions by requiring the model to justify its outputs with biological and clinical features presented in the prompt.

Overall, the detailed prompts act as structured clinical vignettes synthesizing multimodal patient data, enabling downstream LLMs to perform transparent and explainable cardiovascular risk stratification.

\subsection{Fine-Tuning LLMs}
The fine-tuning process leverages the automatically generated Chain-of-Thought (CoT) prompts across all participant tiers to adapt to causal large language models (Causal LLMs) for cardiovascular risk prediction and explanation generation in particular.
 
We fine-tuned the entire dataset of tiers for consistency of information, but for evaluation purposes we also trained on the prompts filtered per tier to facilitate tier-specific training to monitor the LLM's performance. Each prompt-label pair was tokenized separately using tokenizers, then concatenated and padded or truncated to a fixed maximum sequence length ($512$ tokens). Label masking was also applied to ensure the loss is computed only on the target output tokens, preserving prompt tokens as input context.

To ensure robustness, training and evaluation are conducted on a curated subset of $1050$ participants, comprising $350$ participants per tier. A stratified train-test split is employed in prior to the split of subset to preserve balanced label distributions across risk categories. Fine-tuning is performed using parameter-efficient Low-Rank Adaptation (LoRA), with a \texttt{rank} of $8$ and an \texttt{alpha} value of $16$, applied to the attention and feed-forward layers of the transformer. Mixed precision training (FP16) is utilized to accelerate convergence and reduce memory consumption, enabled by NVIDIA A10 GPUs on AWS EC2 instances. 

To quantitatively evaluate model predictions beyond exact text matches, we utilize a semantic similarity metric. Predicted and true labels are embedded using a BioBERT-based SentenceTransformer model \cite{deka2022evidence}, and a cosine similarity threshold of $0.7$ is applied to determine correct semantic matches with the generated outputs. This enables leniency for lexically different yet semantically equivalent diagnoses, providing a more clinically relevant accuracy measure.

Training is performed using the Hugging Face Trainer, which we extend via a custom SemanticTrainer class to incorporate semantic evaluation during validation. Specifically, this semantic check measures the closeness of the predicted diagnosis or risk to the ground-truth label using keyword-level similarity. This approach compensates for the fact that causal language models lack native classification heads and are often sensitive to prompt phrasing. Instead of relying on rigid post-processing or exact string matching, we adopt this more flexible semantic matching strategy since we base this more as a classification task as of now. The training loop includes checkpointing and evaluation at the end of each epoch, with the best-performing model saved based on validation metrics.

\section{Results and Discussion}

This section presents the results obtained from evaluating the three selected base causal LLM models (GPT2\cite{radford2019language}, DeepSeek 1.3B\cite{guo_deepseek-coder_2024}, and Llama 3.2 1B\cite{noauthor_llama_2025}) on the prediction task. The evaluation is conducted based on the accuracy, f1 scores, precision, recall and evaluation loss metrics based on the semantic similarity to the actual classified label. All models were fine-tuned using LoRA-based parameter-efficient fine-tuning (PEFT)~\cite{peft}, with generation restricted to $512$ tokens per response, and the threshold for semantic similarity is set to $0.7$ to tighten the similarity of words and avoid unreliable diagnosis. 

\subsection{Performance}

As mentioned previously, we evaluated the performance of each of the models on accuracy, precision, recall and f1-scores, both on its overall performance and its performance on individual tiers. Tables~\ref{tab:model_performance_overall}, \ref{tab:model_performance_tier1}, \ref{tab:model_performance_tier2} and \ref{tab:model_performance_tier3} display these results. 




\begin{table}[ht]
\centering
\caption{Overall Performance Comparison of LLMs}
\label{tab:model_performance_overall}
\begin{tabular}{lcccc}
\toprule
\textbf{Model} & \textbf{Acc }& \textbf{Prec }& \textbf{Rec} & \textbf{F1} \\
\midrule
GPT-2         & 0.800 & 0.810 & 0.809 & 0.810 \\
LLaMA-3.2 1B  & 0.901 & 0.832 & 0.780 & 0.790 \\
\textbf{DeepSeek 1.3B }&\textbf{ 0.910} & \textbf{0.869 }& \textbf{0.830 }&\textbf{ 0.840} \\
\bottomrule
\end{tabular}
\end{table}

\begin{table}[ht]
\centering
\caption{Performance on Tier 1 Participants}
\label{tab:model_performance_tier1}
\begin{tabular}{lcccc}
\toprule
\textbf{Model} & \textbf{Acc} & \textbf{Prec} & \textbf{Rec} & \textbf{F1} \\
\midrule
GPT-2         & 0.810 & 0.822 & 0.840 & 0.830 \\
LLaMA-3.2 1B    & 0.920 & 0.830 & 0.891 & 0.840 \\
\textbf{DeepSeek 1.3B} & \textbf{0.920} & \textbf{0.831} & \textbf{0.810} & \textbf{0.820} \\
\bottomrule
\end{tabular}
\end{table}

\begin{table}[ht]
\centering
\caption{Performance on Tier 2 Participants}
\label{tab:model_performance_tier2}
\begin{tabular}{lcccc}
\toprule
\textbf{Model} & \textbf{Acc} & \textbf{Prec} & \textbf{Rec} & \textbf{F1} \\
\midrule
GPT-2         & 0.800 & 0.813 & 0.791 & 0.800 \\
LLaMA-3.2 1B    & 0.890 & 0.824 & 0.820 & 0.822 \\
\textbf{DeepSeek 1.3B} & \textbf{0.910} & \textbf{0.850 }&\textbf{ 0.820 }& \textbf{0.830} \\
\bottomrule
\end{tabular}
\end{table}

\begin{table}[ht]
\centering
\caption{Performance on Tier 3 Participants}
\label{tab:model_performance_tier3}
\begin{tabular}{lcccc}
\toprule
\textbf{Model} & \textbf{Acc} &\textbf{ Prec} & \textbf{Rec} & \textbf{F1} \\
\midrule
GPT-2         & 0.811 & 0.890 & 0.812 & 0.842 \\
LLaMA-3.2 1B    & 0.880 & 0.822 & 0.790 & 0.790 \\
\textbf{DeepSeek 1.3B } & \textbf{0.892 }& \textbf{0.860} & \textbf{0.840} & \textbf{0.832} \\
\bottomrule
\end{tabular}
\end{table}

Here, DeepSeek 1.3B achieved the highest performance overall, with an accuracy of $0.910$, precision of $0.869$, recall of $0.830$, and F1 score of $0.840$. This suggests that DeepSeek is the most effective in correctly classifying samples while maintaining a balanced trade-off between precision and recall. This is followed by Llama 3.2 1B that achieves a $0.901$ accuracy and a moderately lower recall of $0.780$, which slightly lowered its F1 score to $0.790$. Finally, GPT-2 performs reasonably well with a F1 score of $0.810$, but is lagged behind the more recent and larger models.

\begin{figure}[ht]
    \centering
    \includegraphics[width=0.5\textwidth]{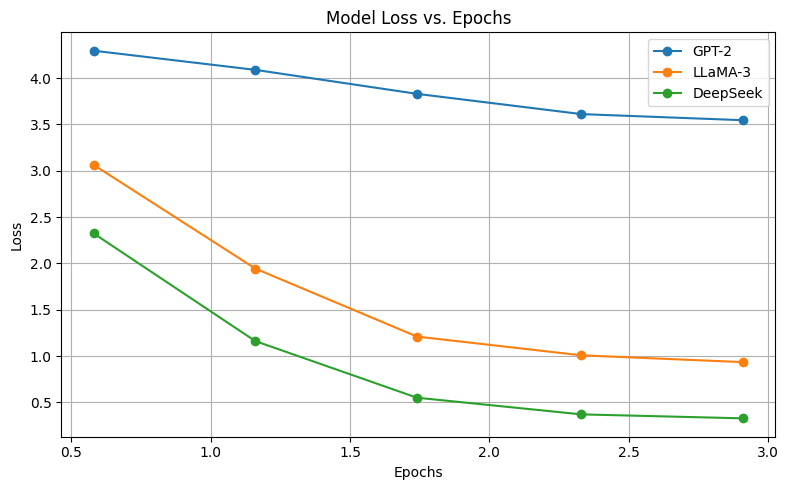}
    \caption{Epoch-wise training loss comparison accross three models.}
    \label{fig:loss-per-training-step}
\end{figure}

Furthermore, the figure \ref{fig:loss-per-training-step} compares the training loss curves of these fine-tuned models across approximately 3 epochs. The DeepSeek fine-tuning exhibits the most rapid convergence, achieving a loss below $0.5$ by epoch $\approx1.75$. In contrast, Llama 3.2 1B shows moderate convergence, while GPT-2 converges the slowest, with a final loss of $3.546$, consistent with the nature of the models that has been previously exhibited.

This trend suggests that model architecture and pretraining quality significantly influence the efficiency of adaptation in fine-tuning. DeepSeek’s fast convergence may indicate better initialization or pretraining alignment with the task data. Meanwhile, GPT-2, being a smaller and older architecture, shows slower adaptation and likely requires either more epochs or lower learning rates to match performance.

Additionally, the shape of each curve provides insight into optimization stability. All three models exhibit monotonic decreases in loss, indicating stable training with no signs of overfitting within the observed window. However, DeepSeek’s steep initial drop followed by a plateau suggests that it reaches saturation quicker, while GPT-2 still shows room for improvement beyond 3 epochs.

\subsection{Tier-wise Analysis}

To further dissect the models' behaviors, we examined performance across participant tiers (Tier 1, 2, and 3), each representing a subset with distinct data characteristics or difficulty levels.

\subsubsection{Tier 1 Participants}

As shown in Table~\ref{tab:model_performance_tier1}, DeepSeek and Llama-3.2 1B performed equally well in terms of accuracy ($0.920$). However, their F1 scores reveal subtle differences. Llama-3.2 1B had a slightly higher recall ($0.891$ vs. $0.810$), indicating it was more effective at retrieving positive cases in this tier, while DeepSeek had more balanced precision and recall, leading to a comparable F1 of $0.820$. GPT-2, though consistent, showed slightly lower performance overall, with a F1 score of $0.830$.

\subsubsection{Tier 2 Participants}

Table~\ref{tab:model_performance_tier2} shows that DeepSeek1.3B again led with a $0.910$ accuracy and a F1 score of $0.830$, confirming its robustness across different data distributions. Llama-3.2 1B was close behind, while GPT-2 maintained a steady but lower performance (F1 of $0.800$). The margins here suggest that newer models generalize better to moderately challenging data.

\subsubsection{Tier 3 Participants}

Tier 3 results (Table~\ref{tab:model_performance_tier3}) indicates a stronger performance by DeepSeek1.3B in handling the most challenging or noisy subset. DeepSeek attained a F1 score of $0.83$, with consistent precision ($0.86$) and recall ($0.84$). Llama-3.2 1B showed a noticeable drop in recall ($0.79$), suggesting difficulty in capturing all relevant cases, which led to a slightly lower F1 score. GPT-2 surprisingly performed better here (F1 of $0.84$) than in Tier 2, likely due to domain-specific overfitting or coincidental alignment with simpler patterns in Tier 3 data.

\subsection{Discussion}

Overall, DeepSeek 1.3B consistently outperformed the other models across all tiers and metrics, showcasing its strength in generalization and balanced predictions when presented with carefully constructed chain-of-thought prompts. Llama-3.2 1B demonstrated competitive performance but occasionally suffered from lower recall, especially in Tier 2 where there were some cardiac labels but not all, and it heavily required inference from ECG embeddings and genotype cluster associations. GPT-2, while still viable, exhibited noticeable limitations due to its smaller architecture and older design.

These results underscore the value of using larger and more recent models for downstream biomedical or genotype-based classification tasks. Tier-wise analysis further emphasizes that no single metric can fully capture model behavior, and precision–recall trade-offs are particularly important when interpreting model effectiveness on edge-case subgroups.

In addition to evaluation metrics, we tracked the training loss progression across epochs to assess convergence behavior. DeepSeek exhibited rapid and consistent convergence, Llama-3.2 1B showed a slightly slower descent, and GPT-2, showed a more gradual and higher loss trajectory. These patterns reflect the models’ respective capacities to absorb the training signal and adapt to the prompt–response structure of the classification task. Furthermore, the learning rate remained constant across training, affirming that improvements were due to architectural and representational strength rather than aggressive optimization.

These insights suggest that DeepSeek 1.3B not only produces better downstream predictions but also learns more efficiently, potentially due to architectural advantages and a better inductive bias for biomedical text and reasoning.

\section{Conclusion}

This study presents a pipeline for interpretable cardiac risk stratification using parameter-efficient fine-tuning of decoder-only large language models (LLMs) on structured genomic data. We introduced a three-tier system to capture varying levels of disease certainty and genetic linkage, where Tier 1 includes directly associated SNP–cardiac condition pairs, Tier 2 involves weaker, semi-curated cardiac associations and Tier 3 uses unsupervised clustering to infer risk for unlabeled participants. Chain-of-Thought (CoT) prompts were generated for each participant regardless of the tier to enable fine-tuning on both diagnostic prediction and justification tasks.

By converting genotype profiles into descriptive, semantically rich prompts, we enabled models such as GPT-2, Llama-3.2 1B, and DeepSeek 1.3B to perform classification and explanation tasks in a clinical genomics context. Experimental results confirm that LLMs can effectively learn from these prompts even at relatively small parameter scales.

Across all tiers, DeepSeek 1.3B consistently outperformed the other models, achieving an overall accuracy of $0.910$, with a F1 score of $0.84$ (Table~\ref{tab:model_performance_overall}). In Tier 1, where the genotype and phenotype link is strongest, both Llama-3.2 1B and DeepSeek reached $0.910$ accuracy, suggesting that current LLMs can internalize well-defined genetic-disease associations through CoT learning. Notably, DeepSeek maintained robust performance even in Tier 3, which lacks direct supervision, achieving $0.892$ accuracy and an F1-score of $0.832$. This highlights its ability to generalize from CoT reasoning derived from unsupervised clustering.

These findings demonstrate that prompt-based LLMs can learn latent structure from genomic data and apply it to novel, ambiguous cases. GPT-2, while less performant ($0.80-0.81$ accuracy across tiers), served as a strong baseline and highlighted the limitations of smaller-capacity models without instruction tuning.

Despite the encouraging results, our approach is subject to several limitations. Most notably, training and evaluation were conducted on a relatively small cohort of $1,050$ participants, with only $350$ individuals allocated per tier. This limited sample size reduces the statistical power and may affect the generalizability of the models. Future work that will be conducted under improved resource conditions could address this constraint appropriately. While we deliberately employed smaller language models to assess feasibility, scaling to larger architectures such as Llama-3 7B\cite{llama3modelcard} or GPT-4\cite{gpt4} could substantially enhance performance, particularly in Tier 3 scenarios or zero-shot classification tasks.

Furthermore, as causal language models do not incorporate native classification heads, their outputs are often sensitive to prompt structure and may lack consistency in diagnostic phrasing. To address this, semantic similarity metrics were employed to align generated outputs with known cardiac conditions. While classical NLG metrics such as BLEU, METEOR, or ROUGE-SEM~\cite{zhang_rouge-sem_2024} were not designed for medical classification tasks, they will be adapted here to quantitatively assess tier-based predictive performance. Moving forward, post-processing strategies will be incorporated to constrain output variability and enforce label consistency, particularly important in clinical forecasting tasks, where the interpretation of subclinical risk must remain robust and standardized across decoders like Llama.

Moreover, utilizing techniques such as knowledge graphs will be a significant focus in the future work to capture significant relationships that clusters may not be able to capture. Knowledge graphs \cite{xu_knowledge_2024} can help capture and model relationships between diseases, symptoms, treatments, and other medical entities in a structured form, allowing the model to reason more effectively and improve decision-making. Integrating richer context from electronic health records (EHRs), especially concerning blood test reports and diet-based information could unlock more holistic reasoning capabilities, enabling the model to provide more personalized and comprehensive recommendations. Incorporating multi-modal data from medical imaging, genomics, and lab results could further enhance the model’s diagnostic abilities, making it a more powerful tool for clinicians.

Overall, our findings suggest that natural language interfaces, powered by LLMs, offer a viable pathway toward interpretable and flexible clinical decision support systems. These systems could potentially improve clinical workflows by providing accurate, context-aware insights while ensuring the transparency and interpretability of AI-driven decisions. With further refinement and expansion, these systems could become indispensable tools in healthcare, supporting clinicians in making more informed and timely decisions.

\bibliographystyle{IEEEtran}
\bibliography{acl2016}

\end{document}